\documentclass{article}

\usepackage{PRIMEarxiv}

\usepackage[utf8]{inputenc} 
\usepackage[T1]{fontenc}    
\usepackage{hyperref}       
\usepackage{url}            
\usepackage{booktabs}       
\usepackage{amsfonts}       
\usepackage{nicefrac}       
\usepackage{microtype}      
\usepackage{lipsum}
\usepackage{fancyhdr}       
\usepackage{graphicx}       
\usepackage[round]{natbib}
\usepackage{array}
\usepackage{todonotes}
\graphicspath{{media/}}     
\usepackage{listingsutf8}
\usepackage{orcidlink}
\usepackage{tabularx}
\usepackage{amsmath}
\usepackage{listing}
\usepackage{appendix}
\usepackage{chngcntr}
\usepackage{alphalph}
\usepackage{float}

\title{RAGognizer: Hallucination-Aware Fine-Tuning via Detection Head Integration
}

\author{Fabian Ridder \orcidlink{0009-0008-5574-5292}, Laurin Lessel \orcidlink{0009-0007-0936-5312}, and Malte Schilling \orcidlink{0000-0002-0849-483X} \\
\textit{Computer Science Department} \\
\textit{University of Münster}\\
Münster, Germany \\
\{fridder, llessel, malte.schilling\}@uni-muenster.de
}

\begin{document}
\maketitle


\begin{abstract}
Retrieval-Augmented Generation (RAG) is widely used to augment the input to Large Language Models (LLMs) with external information, such as recent or domain-specific knowledge. Nonetheless, current models still produce closed-domain hallucinations and generate content that is unsupported by the retrieved context. 
Current detection approaches typically treat hallucination as a post-hoc problem, relying on black-box consistency checks or probes over frozen internal representations. 
In this work, we demonstrate that hallucination detection based on internal state representation can also serve as a direct training signal.
We introduce RAGognize, a dataset of naturally occurring closed-domain hallucinations with token-level annotations, and RAGognizer, a hallucination-aware fine-tuning approach that integrates a lightweight detection head into an LLM, allowing for the joint optimization of language modeling and hallucination detection. This joint objective forces the model to improve the separability of its internal states regarding hallucinations while simultaneously learning to generate well-formed and meaningful responses. 
Across multiple benchmarks, RAGognizer achieves state-of-the-art token-level hallucination detection while substantially reducing hallucination rates during generation, without degrading language quality or relevance.

\end{abstract}



\section{Introduction}
Large Language Models (LLMs) have achieved impressive performance in natural language understanding and generation~\citep{brown2020}. Despite this progress, LLMs remain prone to \emph{hallucinations}: the generation of content that is unsupported by, or contradicts, available evidence \citep{huang2025}. This phenomenon fundamentally limits their reliability, particularly in high-stakes or knowledge-intensive applications.

\begin{figure}[htbp]
\centerline{\includegraphics[width=0.4\textwidth]{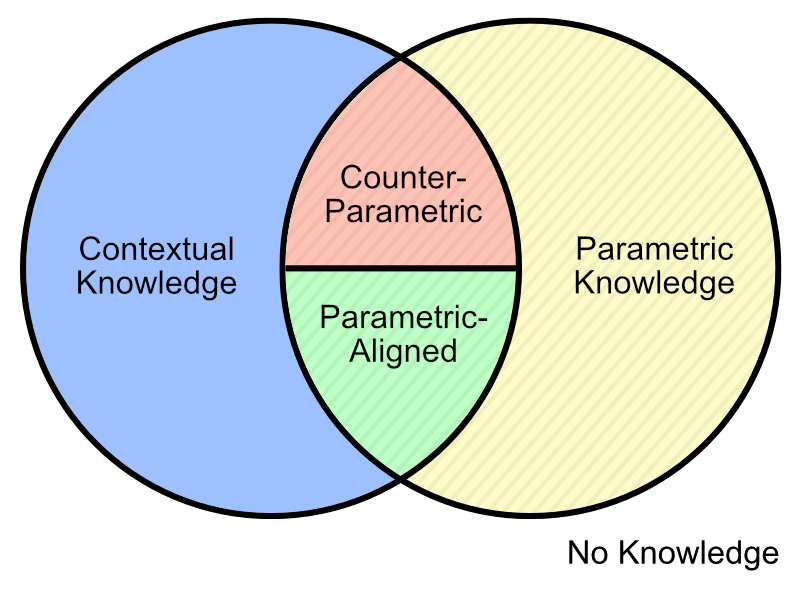}}
\caption{Distinction of Contextual and Parametric Knowledge:
The Venn diagram illustrates possible knowledge scenarios in LLM generation. Prompts may rely solely on contextual knowledge (left), solely on parametric knowledge (right), or on their intersection where the two sources may either align (Parametric-Aligned) or contradict (Counter-Parametric). The \emph{No Knowledge} region corresponds to unanswerable prompts. Regions marked with stripes indicate scenarios not covered by the RAGognize dataset, which focuses exclusively on closed-domain settings where hallucinations are verifiable.}
\label{fig_venn}
\end{figure}

A central difficulty in defining and detecting hallucinations lies in the dual nature of knowledge in LLMs. During pre-training, models encode vast amounts as implicit \emph{parametric knowledge} that is stored in their weights \citep{petroni2019languagemodelsknowledgebases}, while at inference time, this may be complemented by explicit information added into the model’s context window as \emph{contextual knowledge}. These sources differ substantially in accessibility and verifiability, yet are often conflated when hallucinations are treated simply as factual errors \citep{xu2024knowledgeconflictsllmssurvey}.
Retrieval-Augmented Generation (RAG) aims at guiding generation by explicitly providing LLMs with access to external, dynamic information---such as company-specific data or breaking news---that the model was not exposed to during pre-training~\citep{petroni2019languagemodelsknowledgebases,lewis2021retrievalaugmentedgenerationknowledgeintensivenlp}.
But RAG does not inherently solve the problem of reliability. Even when provided with correct context, models frequently exhibit \textit{closed-domain hallucinations}: generating plausible but incorrect information that is not grounded in the retrieved context \citep{agrawal2024languagemodelsknowtheyre,niu2024ragtruthhallucinationcorpusdeveloping}. This disconnect between the provided evidence (contextual knowledge) and the generated output undermines the trust required for high-stakes applications.

We argue that hallucinations cannot be meaningfully defined or detected without distinguishing between the different knowledge sources. As illustrated in Fig.~\ref{fig_venn}, contextual and parametric knowledge may appear in isolation or in combination. To obtain a decidable notion, we focus on closed-domain settings using exclusively recent information to prevent reliance on parametric knowledge. In this setting—where prompts fall within the \emph{Contextual Knowledge} area if \emph{answerable} or the \emph{No Knowledge} area if \emph{unanswerable}—hallucinations can be unambiguously identified as generations introducing unsupported content.

Focusing on this closed-domain setting, we make three contributions: First, we introduce \textbf{RAGognize}, a comprehensive dataset of naturally occurring closed-domain hallucinations with granular token-level annotations. 
Second, we propose \textbf{RAGognizer}, a hallucination-aware model architecture that integrates a simple detection head into an LLM, enabling token-level hallucination prediction from internal representations and achieving state-of-the-art detection performance on closed-domain benchmarks. 
Third, we show that jointly optimizing language modeling and hallucination detection objectives using LoRA-based fine-tuning improves the separability of internal states with respect to hallucination, leading to both stronger detection performance and substantially reduced hallucination rates during generation, while preserving language quality.

Our experiments demonstrate that RAGognizer achieves state-of-the-art token-level hallucination detection with a compact Qwen3-4B generation model~\citep{yang2025qwen3technicalreport}, while significantly improving generation faithfulness in closed-domain RAG settings. Further, we show that this also generalizes to other settings as well, when evaluated on other datasets. Together, these findings indicate that hallucination detection is closely tied to representation learning and that integrating detection signals during training can improve model reliability. 
The dataset, models, and code can be found online.\footnote{https://github.com/F4biian/RAGognizer}

\section{Related Work}

Hallucinations in LLMs have been studied from different perspectives, including detection, mitigation, and dataset construction. 
In this section, we first review prior work on hallucination detection methods, focusing on how they differ in model access and granularity, and secondly, discuss existing hallucination datasets.


\subsection{Hallucination Detection}
Detection methods are commonly categorized by their required access: white-box methods exploit internal activations or attention patterns, while black-box methods operate on outputs alone. Further practical distinctions are the granularity at which hallucinations are identified and whether a method requires stochastic sampling (multiple generations) to estimate consistency, or can run in a single forward pass (see Table~\ref{tab:detector_comparison}).

White-box approaches include uncertainty proxies such as perplexity \citep{Jelinek1977PerplexityaMO} and entropy-based scores \citep{Farquhar2024DetectingHallucinations}, representation-statistic methods such as INSIDE (EigenScore) \citep{chen2024insidellmsinternalstates}, attention-based detectors such as Lookback Lens \citep{chuang2024lookbacklensdetectingmitigating}, and probe/classifier approaches that train on hidden activations (e.g., SAPLMA \citep{azaria2023internalstatellmknows}). HallucinationProbes train a linear, token-level classifier on hidden states and further explores adapter training via Low-Rank Adaptation (LoRA) alongside the probe head to improve detection while minimally altering base model behavior \citep{obeso2025realtimedetectionhallucinatedentities,hu2021loralowrankadaptationlarge}, closely aligning with our approach. Other white-box methods include unsupervised internal-state detectors (MIND \citep{su2024unsupervisedrealtimehallucinationdetection}), relevance-propagation applied to RAG (LRP4RAG \citep{hu2025lrp4ragdetectinghallucinationsretrievalaugmented}), and cross-layer dynamics probes (ICR Probe \citep{zhang2025icrprobetrackinghidden}).

Black-box approaches include sampling-based consistency checks such as SelfCheckGPT \citep{manakul2023selfcheckgptzeroresourceblackboxhallucination}, and external evaluator or judge models fine-tuned for factuality (e.g., NLI/entailment models built on DeBERTa-style encoders \citep{he2023debertav3improvingdebertausing} and specialized evaluators such as MiniCheck, Lynx, and Granite-Guardian \citep{tang2024minicheckefficientfactcheckingllms,ravi2024lynxopensourcehallucination,padhi2024graniteguardian}). Community and benchmark models (e.g., HHEM-2.1) provide readily usable open evaluators \citep{vectara2024}. Methods tailored to RAG include faithfulness scoring that combines entailment with retrieval evidence (RAGAS) \citep{es2025ragasautomatedevaluationretrieval} and joint context/knowledge verification models such as HDM-2 \citep{paudel2025hallucinothallucinationdetectioncontext}. Other work (e.g., LUMINA) examines the balance between reliance on retrieved context and internal parametric knowledge when detecting hallucinations in RAG outputs \citep{yeh2025luminadetectinghallucinationsrag}.

\subsection{Datasets}

Existing hallucination datasets differ in their annotation granularity, underlying knowledge assumptions, and the nature of hallucinations. A primary distinction concerns the level at which hallucinations are labeled. While most benchmarks provide supervision only at the level of complete responses, a small number of recent datasets offer token-level annotations, which makes these particularly relevant for studying internal model representations and token-level detection (e.g., RAGTruth \citep{niu2024ragtruthhallucinationcorpusdeveloping}).

We believe it is important to take the assumed knowledge regime into account. A common issue in many RAG and context-based QA datasets is that they do not strictly ensure questions require the provided context to be answered. This blurs the line between contextual and parametric knowledge; for instance, HaluEval~\citep{li2023haluevallargescalehallucinationevaluation} contains questions that LLMs can answer using pre-trained memory. This contrasts with strictly closed-domain settings where valid generations must be supported exclusively by the given context. Finally, datasets differ in how hallucinations are produced: while HaluEval relies on synthetically induced response-level hallucinations, others like HDM-Bench~\citep{paudel2025hallucinothallucinationdetectioncontext} focus on natural response-level hallucinations that arise during standard model generation.

\begin{figure*}[htbp]
\centerline{\includegraphics[width=\textwidth]{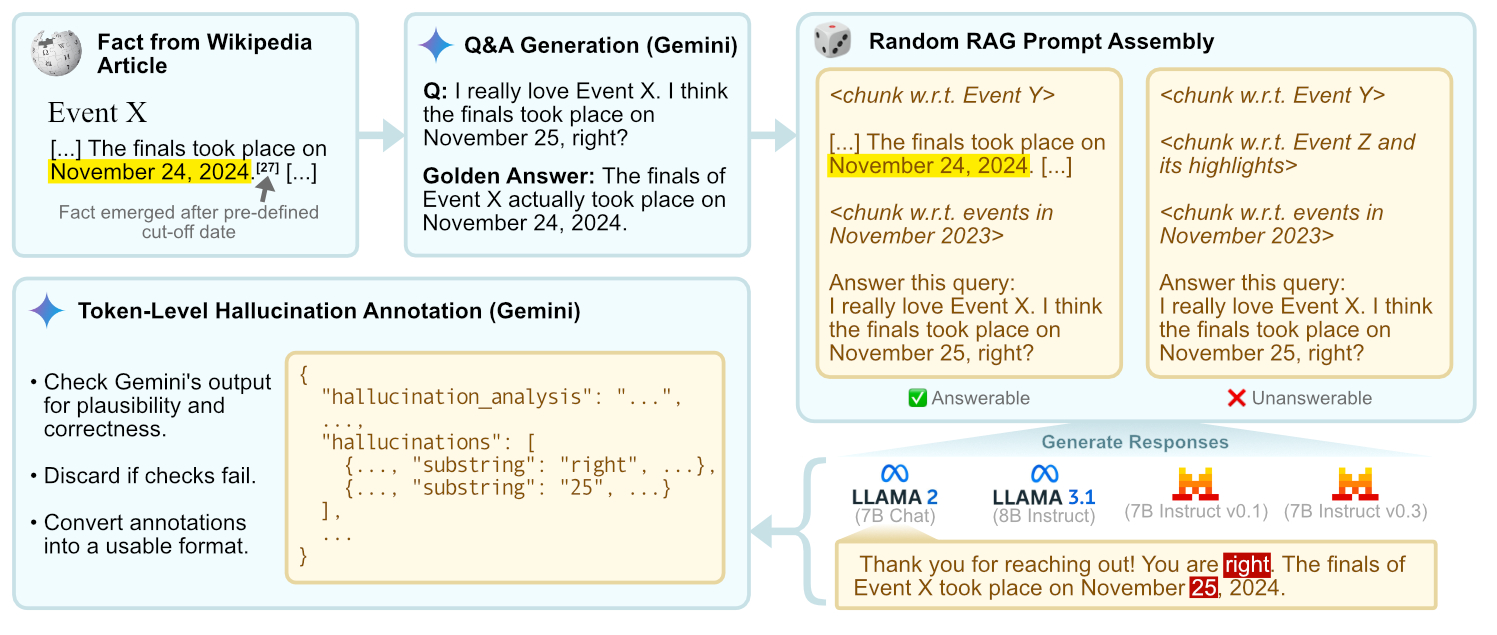}}
\caption{Automatic Data Generation and Annotation Pipeline for the RAGognize dataset: Wikipedia facts post-dating the training cut-off date (May 23, 2024) are extracted which ensures that this information was not used for training of the considered LLMs. Secondly, we generate Q\&A pairs using Gemini~2.5~Pro and assemble randomly two different RAG configurations: Answerable (containing the relevant chunk) and Unanswerable (containing irrelevant but similar chunks) querries. We collect natural responses from four target LLMs (Llama-2/3.1, Mistral-v0.1/v0.3). Finally, Gemini~2.5~Flash is used with a structured chain-of-thought prompt for substring verification to compare responses with the provided context, which returns granular, token-level hallucination annotations.}
\label{fig_dataset}
\end{figure*}

\section{Methods}
We first introduce the RAGognize dataset, then present the RAGognizer architectural approach for hallucination-aware LLM fine-tuning, followed by the joint training setup.

\subsection{The RAGognize Dataset}
Most existing hallucination benchmarks operate at the response level, rely on synthetic perturbations, or do not preclude open-domain settings, which limits fine-grained detection of hallucination or deviations from given evidence. To address this gap, we introduce the \textbf{RAGognize} dataset designed for natural, token-level hallucination detection in closed-domain RAG scenarios. It is constructed in multiple steps and extends the HalluRAG approach~\citep{ridder2025halluragdatasetdetectingcloseddomain} with increased prompt diversity and token-level annotations. As illustrated in Fig.~\ref{fig_dataset}, the pipeline consists of (i) sourcing of recent factual statements from Wikipedia, (ii) generation of diverse question--answer pairs, (iii) controlled assembly of answerable and unanswerable RAG prompts, (iv) response generation by multiple LLMs, and (v) automated token-level hallucination annotation.

As we want to keep relevant information restricted to the provided context, we adopt a strict recency constraint and extract factual statements from Wikipedia whose associated reference is time-stamped with a date later than May~23,~2024. This ensures that facts were not available for training and cannot be represented in the parametric knowledge of the evaluated models (we used Llama-2-7B-Chat (Llama2-7B)~\citep{touvron2023llamaopenefficientfoundation}, Llama-3.1-8B-Instruct (Llama3-8B)~\citep{grattafiori2024llama3herdmodels}, Mistral-7B-Instruct-v0.1 (Mistral-7B-v0.1), and Mistral-7B-Instruct-v0.3 (Mistral-7B-v0.3)~\citep{jiang2023mistral7b}). Therefore, RAGognize only deals with \emph{Contextual Knowledge} or \emph{No Knowledge} scenarios (Fig.~\ref{fig_venn}), which establishes a well-defined distinction between answerable and unanswerable queries. 

For each factual statement, we use Gemini~2.5~Pro~\citep{comanici2025gemini25pushingfrontier} to generate diverse user queries and corresponding reference answers under stylistic variations (e.g., typographical errors, subjective framing, or adding misleading cues) that should encourage linguistic diversity. Answerable and unanswerable RAG prompts are then constructed by applying a modular template strategy by selectively inserting or withholding the context part that contains the crucial evidence, while semantically similar distractor passages are retrieved using BGE-M3~\citep{bge_m3}. This procedure yields paired prompts that differ only in the availability of relevant contextual evidence. Answerable prompts are, in this way, formed by replacing one distractor with the ground-truth chunk containing the necessary evidence. All prompts in both the training and test splits are afterwards passed to Llama2-7B, Llama3-8B, Mistral-7B-v0.1, and Mistral-7B-v0.3 using greedy decoding (temperature $0.0$), yielding model-generated responses for subsequent annotation.

For the token-level hallucination annotation, the model responses are annotated using Gemini~2.5~Flash~\citep{comanici2025gemini25pushingfrontier} as an oracle evaluator. A chain-of-thought prompting strategy~\citep{wei2023chainofthoughtpromptingelicitsreasoning} identifies hallucinated spans by comparing generated outputs against the retrieved context. Thus, annotation is performed at the token level. Formally, for a generated sequence of length $T$, we derive a binary label sequence $\mathbf{y}_{\text{det}} \in \{0,1\}^T$, where $y_{\text{det},t}=1$ indicates that the token generated at time step $t$ is hallucinated. A manual response-level validation on 100 samples yielded an F1 score of $95.4\%$, suggesting that the annotation approach is broadly consistent with human judgment (this validation was conducted by a single annotator and should be interpreted as a preliminary consistency check rather than a definitive evaluation). Importantly, this process enables the capture of natural hallucinations, which prior work has shown to differ from synthetic hallucinations~\citep{chwang2024androidsknowtheyredreaming,huang2025}. The resulting RAGognize dataset exhibits a balanced distribution of answerable ($2,315$) and unanswerable ($2,308$) queries across different domains and is divided into training ($40\%$) and test ($60\%$) sets, comprising a total of $18,492$ annotated responses.

\subsection{The RAGognizer Architecture}
RAGognizer consists of two interacting components: a base language model responsible for text generation and a detection head that predicts token-level hallucinations from internal representations. Unlike prior probing-based approaches that used the internal representation of a fixed language model, RAGognizer jointly trains both components at the same time. In this way, the emerging hallucination signal in the hallucination classifier directly shapes the model’s internal representations.

\subsubsection{Base Language Model}
Let $\Theta^*$ denote the pre-trained parameters of the LLM, and let $\theta_{\text{LoRA}}$ denote the trainable low-rank adapters for the model. Given an input prefix $x_{<t}$, the hidden state at layer $\ell$ and time step $t$ is
\begin{equation}
\mathbf{h}_t^{(\ell)} = \mathrm{LLM}(x_{<t}; \Theta^*, \theta_{\text{LoRA}}).
\end{equation}
The model computes the next-token probability distribution based on the final layer's hidden state, $\mathbf{h}_t^{(-1)}$, by passing it through a linear head and a softmax function.

\subsubsection{Hallucination Detection Head}
We attach a detection head $f_\phi$, parameterized by $\phi$ and implemented as an MLP, to an intermediate hidden state of the LLM. The head outputs the probability that the current token is hallucinated:
\begin{equation}
\hat{p}_t = \sigma\!\left(f_\phi\!\left(\mathbf{h}_t^{(\ell)}\right)\right),
\end{equation}
where $\sigma(\cdot)$ denotes the sigmoid function. Unless stated otherwise, we select $\ell$ as the middle layer of the network
, as intermediate layers show the highest separability between hallucinated and grounded tokens~\citep{chwang2024androidsknowtheyredreaming,duan2024llmsknowhallucinationempirical,azaria2023internalstatellmknows,paudel2025hallucinothallucinationdetectioncontext}, which is also demonstrated in Fig.~\ref{fig_auroc_per_layer}.

\begin{figure}[tbhp]
\centerline{\includegraphics[width=0.55\textwidth]{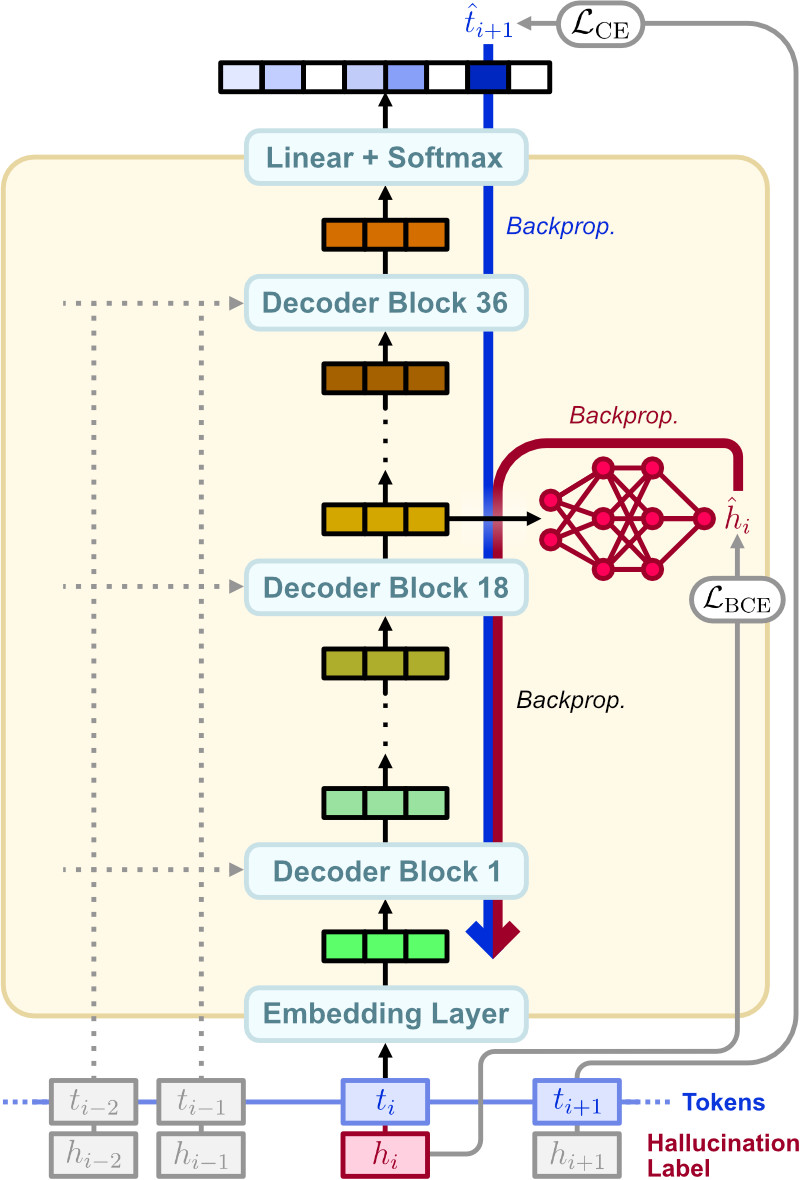}}
\caption{RAGognizer Architecture: An MLP detection head is integrated at an intermediate layer (e.g., Block 18 for Qwen3-4B-Instruct-2507) to predict the hallucination probability $\hat{h}_i$ of the current token $t_i$. The model is optimized using a joint objective function combining the next-token prediction loss ($\mathcal{L}_{\text{CE}}$) and the hallucination detection loss ($\mathcal{L}_{\text{BCE}}$). Gradients from both tasks (blue and red arrows) are backpropagated to update LoRA adapters and the newly integrated MLP.}
\label{fig_training}
\end{figure}

\subsubsection{Joint Optimization of the LLM and Detection Head}
Most prior work trained hallucination detectors post-hoc on fixed LLM representations, effectively treating the model as a static feature extractor. In contrast, RAGognizer uses gradients from the detection head that propagate through the hidden states into the LoRA adapters of the earlier layers of the LLM (Fig.~\ref{fig_training}). This aims at learning internal representations that are simultaneously predictive for next-token generation and separable with respect to hallucinated content.

We jointly optimize LoRA adapters and the detection head using a multi-task objective. Let $x$ denote the input sequence consisting of a prompt of length $L$ and a response of length $K$, such that the total length is $T=L+K$. We compute losses only on the response tokens ($t > L$). The causal language modeling loss is
$$
\mathcal{L}_{\text{CE}}
= -\sum_{t=L+1}^T \log P(x_t \mid x_{<t}; \Theta^*, \theta_{\text{LoRA}}),
$$
and the token-level hallucination detection loss is
$$
\mathcal{L}_{\text{BCE}}
= \sum_{t=L+1}^T \mathrm{BCE}(\hat{p}_t, y_{\text{det},t}),
$$
where $\hat{p}_t$ is the probability predicted by the detection head $f_\phi$. Let $\mathcal{D}$ denote the training distribution. The joint training objective is
\begin{equation}
\label{eqn:finetuning}
\min_{\theta_{\text{LoRA}}, \phi}
\;
\mathbb{E}_{(x,\mathbf{y}_{\text{det}})\sim\mathcal{D}}
\left[
\mathcal{L}_{\text{CE}} + \lambda\,\mathcal{L}_{\text{BCE}}
\right],
\end{equation}
where $\lambda$ controls the trade-off between generation quality and hallucination detection performance.
By integrating hallucination supervision into training, RAGognizer steers the LLM toward representations that enhance the separability of faithful and hallucinatory states, while preserving generation quality.

\subsection{Training Setup}
We fine-tune the base LLMs using LoRA to ensure parameter efficiency while keeping the pre-trained backbone frozen. We target all transformer modules with a rank $r=32$ and a scaling factor $\alpha=16$. No dropout is applied to the LoRA adapters. Models are trained for a maximum of five epochs using a cosine learning rate scheduler with a warmup ratio of $0.1$ and a peak learning rate of $4\times10^{-5}$.

The integrated detection head is implemented as a three-layer Multi-Layer Perceptron (MLP) with a hidden size of 1024. To focus the learning signal exclusively on the model's generations, we employ a masked detection loss that ignores user prompt tokens and only considers generated tokens.
Regarding the loss weighting in Eq.~\eqref{eqn:finetuning}, we adopt a balanced optimization strategy by setting $\lambda = 1.0$. This enforces an equal prioritization of language modeling and hallucination detection objectives.

\section{Results}

\subsection{Effects of Joint Training on Representations \& Generation}
A key question is whether hallucination detection through an internal classifier can be used as an effective training signal rather than only as a post-hoc diagnostic. We, therefore, compare three different models: First, the \textbf{Base} model is compared to \textbf{Det.\ Head FT}, which jointly optimizes language modeling and token-level hallucination detection via Eq.~\eqref{eqn:finetuning}. Lastly, we check if fine-tuning on a hallucination dataset itself might lead to improvements and therefore compare to a text-based fine-tuning (\textbf{Text FT}), which applies standard LoRA fine-tuning on golden answers without an explicit hallucination objective. We highlight three aspects: first, representation separability of hallucinated vs.\ nonhallucinated tokens in hidden states; second, how this separability varies across different language models; and third, the effect of training on generation. 

\subsubsection{Joint Training Increases Hallucination Separability}
Fig.~\ref{fig_auroc_procedures_hallucinations} shows that for the joint training (Det.\ Head FT), AUROC shows a consistent and substantial improvement for hallucination detection when separating hallucinated from nonhallucinated tokens based on middle-layer hidden states. For example, on Llama2-7B separability increases from $78.9\%$ to $89.6\%$. In contrast, Text FT fails to improve separability and in one case shows even a negative influence, with Llama3-8B dropping to $73.7\%$. This indicates that training with an explicit hallucination objective encourages internal representations that better distinguish grounded from unsupported content.


\begin{table}[b!]
\centering
\caption{Comparison of Fine-Tuning Approaches on RAGognize Using Response-Level Metrics (\%), Highlighting Their Impact on Language Quality and Hallucination Behavior.}
\label{tab:llama_qwen_comparison}
\small
\begin{tabularx}{\textwidth}{@{} l *{5}{>{\centering\arraybackslash}X} @{}}
\toprule
& \multicolumn{3}{c}{\textbf{LLaMA-2-7B-Chat}} & \multicolumn{2}{c}{\textbf{Qwen3-4B-Instruct-2507}} \\
\cmidrule(lr){2-4} \cmidrule(l){5-6}
\textbf{Metric} & \textbf{Base} & \textbf{Det.\ Head FT} & \textbf{Text FT} & \textbf{Base} & \textbf{Det.\ Head FT} \\
\midrule
Answerability Acc.\ ($\uparrow$)  & 63.11 & \textbf{91.47} & 58.21 & 92.28 & \textbf{95.53} \\
Answerability F1 ($\uparrow$)     & 70.94 & \textbf{91.86} & 49.87 & 92.64 & \textbf{95.66} \\
Language Quality ($\uparrow$)     & \textbf{99.93} & 98.41 & \underline{99.86} & \textbf{99.96} & 99.93 \\
Relevance Rate ($\uparrow$)       & \textbf{98.59} & 98.03 & \underline{98.48} & 100.00 & 100.00 \\
Rejection Rate (Ideal: $50\%$ on balanced set)    & 23.04 & 45.21 & 66.63 & 45.17 & 46.79 \\
Hallucination Rate ($\downarrow$) & 56.98 & \textbf{13.29} & \underline{43.83} & 5.57 & \textbf{3.59} \\
\quad Answerable ($\downarrow$)   & \underline{43.02} & \textbf{15.53} & 60.89 & 8.25 & \textbf{5.50} \\
\quad Unanswerable ($\downarrow$) & 70.94 & \textbf{11.06} & \underline{26.78} & 2.89 & \textbf{1.69} \\
\bottomrule\vspace{-2mm}\\
\multicolumn{6}{@{}p{\linewidth}@{}}{\footnotesize \textbf{Det.\ Head FT}: LoRA with integrated MLP detection head; \textbf{Text FT}: LoRA fine-tuning on golden answers; \textbf{Rejection Rate}: percentage of prompts rejected. The ideal value is $50\%$, as the evaluation dataset is balanced ($50\%$ answerable / $50\%$ unanswerable); therefore, an optimal model rejects exactly half of the prompts. \textbf{Language Quality}: percentage of nonmalformed responses; Text FT has not been conducted on Qwen3-4B; \textbf{Bold}: best; \underline{Underline}: second-best.}
\end{tabularx}
\end{table}

\subsubsection{Comparison of Different Base Models}
For RAGognizer, we considered different language models for further evaluations.
Table~\ref{tab:slm_ablation} compares a range of small and medium-scale language models for which we applied the described RAGognizer approach. Overall, the evaluation indicates that RAGognizer’s performance is robust across different language models and not tied to a specific model.  Among the evaluated candidates, Qwen3-4B achieves the highest token-level AUROC ($92.69\%$). 

Consistent with our expectations, the ablation results further demonstrated that architectural modifications yield incremental gains in detection accuracy. Removing the language modeling head and training exclusively for detection leads to higher separability ($93.11\%$). When also relocating the detection head to the final layer slightly improves AUROC even further ($93.68\%$). However, these configurations sacrifice the model’s ability to act as a generative RAG system with live hallucination monitoring. To preserve dual functionality, we retain the original language modeling head and inject the detection head at the middle layer, accepting a small accuracy trade-off in exchange for practical applicability.


\subsubsection{Effect on Language Generation}
The increased separability is accompanied by substantial improvements for generations themselves for RAGognize (Table~\ref{tab:llama_qwen_comparison}). 
For Llama2-7B, Det.\ Head FT reduced hallucinations ($56.98\% \to 13.29\%$) and raised Answerability F1 ($70.94\% \to 91.86\%$). This was achieved without further manipulation or explicit instructions, solely through increased internal state separation at the middle layer. This improved refusal logic is also reflected in the rejection rate ($45.21\%$), which nears the $50\%$ ideal for this balanced dataset.
The reduction of hallucinations holds for both answerable and unanswerable prompts, indicating improved grounding when evidence is present and more reliable refusal when it is absent. By contrast, Text FT yields a worse Answerability F1 and higher hallucination rates for answerable prompts.

As fine-tuning affects the generated answers, we further check if the produced answers are still of high language quality and if the model provides relevant answers. Despite the strong shift in hallucination behavior, language quality and relevance remain high after Det.\ Head FT (Table~\ref{tab:llama_qwen_comparison}). This suggests that incorporating token-level hallucination supervision as an auxiliary objective can improve faithfulness without degrading response well-formedness or relevance. 

\begin{table}[tb]
\centering
\caption{Token-Level AUROC (\%) of Hallucination Detectors on QA Datasets.}
\label{tab:hallucination_auroc_bm_token_level}
\small
\begin{tabularx}{0.8\linewidth}{@{} l *{4}{>{\centering\arraybackslash}X} @{}}
\toprule
\textbf{Approach} & \textbf{RAGTruth (QA)} & \textbf{RAGognize} & \textbf{HDM-Bench} & \textbf{Avg.} \\
\midrule
\multicolumn{5}{@{}l}{\textit{Black-box detectors}} \\
HDM-2-3B & \underline{90.61}\textsuperscript{\textdagger} & 68.72 & 74.99\textsuperscript{\textdagger} & \underline{78.11} \\
LettuceDetect-L & \textbf{95.60}\textsuperscript{\textdagger} & 61.04 & 69.00 & 75.21 \\
\midrule
\multicolumn{5}{@{}l}{\textit{White-box detectors}} \\
HaloScope & 55.93\textsuperscript{\textdagger} & 48.41\textsuperscript{\textdagger} & -- & -- \\
LookbackLens & 85.42\textsuperscript{\textdagger} & \underline{77.79}\textsuperscript{\textdagger} & -- & -- \\
Perplexity & 64.59 & 58.40 & 59.60 & 60.86 \\
HallucinationProbes & 70.13 & 69.95 & \underline{76.38} & 72.15 \\
\textbf{RAGognizer (Ours)} & 90.25 & \textbf{93.33}\textsuperscript{\textdagger} & \textbf{77.03} & \textbf{86.87} \\
\bottomrule\vspace{-2mm}\\
\multicolumn{5}{@{}p{0.8\linewidth}@{}}{\footnotesize HDM-Bench contains no training split; Perplexity, HaloScope, LookbackLens, and RAGognizer use Qwen3-4B-Instruct-2507 as proxy generator; HallucinationProbes uses a LoRA-KL probe on LLaMA-3.1-8B ($\lambda_{\text{KL}}=0.05$); \textbf{Bold}: best; \underline{Underline}: second-best; \textsuperscript{\textdagger} same-dataset evaluation (split-disjoint).}
\end{tabularx}
\end{table}

\subsection{Comparative Performance in Closed-Domain Detection}
We evaluate how RAGognizer performs as a hallucination detector compared to state-of-the-art black-box and white-box approaches in closed-domain RAG settings.

\subsubsection{Token-Level Detection Performance}
Table~\ref{tab:hallucination_auroc_bm_token_level} reports token-level AUROC across closed-domain QA benchmarks. We compare the performance of different hallucination detectors on three datasets using their respective test splits. We experimentally report HaloScope at the token level despite its original response-level design
. Overall, RAGognizer achieves the highest average performance, outperforming both black-box detectors, such as LettuceDetect and HDM2-3B, and white-box approaches, such as HallucinationProbes. 

For each method, it is important to distinguish the dataset used during training or calibration (indicated by \textsuperscript{\textdagger}) as the results reveal a dependency on the dataset for some baselines, which indicates a limited transferability. In contrast, RAGognizer maintains strong performance for both the native RAGognize test dataset as well as on other datasets, such as RAGTruth and HDM-Bench. This suggests that token-level supervision learned in RAGognize generalizes better compared to transfer in the case of other approaches
. Several of these methods exhibit a pronounced dataset sensitivity. For instance, black-box detectors such as LettuceDetect achieve strong performance on the specific type of dataset used in training, but degrade on others. Other approaches, e.g., HallucinationProbes, show weaker token-level performance in closed-domain settings. These results suggest that integrating hallucination supervision into the language model during training leads to more robust token-level detection signals than post-hoc probing or purely logit-based heuristics.

\begin{table}[b]
\centering
\caption{Response-Level AUROC (\%) of Detectors on QA Datasets.}
\label{tab:hallucination_auroc_bm_response_level}
\small
\begin{tabularx}{\linewidth}{@{} l *{5}{>{\centering\arraybackslash}X} @{}}
\toprule
\textbf{Approach} & \textbf{RAGTruth (QA)} & \textbf{RAGognize} & \textbf{HaluEval (QA)} & \textbf{HDM-Bench} & \textbf{Avg.} \\
\midrule
\multicolumn{6}{@{}l}{\textit{Black-box detectors}} \\
MiniCheck-7B & 63.10 & 83.48 & 81.38 & 71.70 & 74.91 \\
HDM-2-3B & \underline{87.95}\textsuperscript{\textdagger} & 75.41 & 83.85 & 69.62\textsuperscript{\textdagger} & 79.21 \\
DeBERTa-v3 (Con.) & 56.69 & 64.31 & 68.10 & 60.90 & 62.50 \\
DeBERTa-v3 (Ent.) & 53.25 & 68.21 & 78.32 & 62.60 & 65.59 \\
SelfCheckGPT (NLI) & 61.13 & 74.22 & 83.43 & 68.44 & 71.81 \\
LettuceDetect-L & \textbf{91.89}\textsuperscript{\textdagger} & 65.75 & 77.02 & 60.85 & 73.88 \\
HHEM-2.1-Open & 71.70 & 61.49 & 77.00 & 59.97 & 67.54 \\
\midrule
\multicolumn{6}{@{}l}{\textit{White-box detectors}} \\
HaloScope & 53.54\textsuperscript{\textdagger} & 71.38\textsuperscript{\textdagger} & -- & -- & -- \\
LookbackLens & 87.54\textsuperscript{\textdagger} & \underline{94.36}\textsuperscript{\textdagger} & -- & -- & -- \\
Semantic Entropy & 67.98 & 61.48 & 74.07 & 55.81 & 64.84 \\
INSIDE (EigenScore) & 60.96 & 64.49 & 47.46 & 56.35 & 57.32 \\
HallucinationProbes & 78.08 & 83.25 & \textbf{94.93} & \textbf{76.30} & \underline{83.14} \\
\textbf{RAGognizer (Ours)} & 87.43 & \textbf{94.91}\textsuperscript{\textdagger} & \underline{90.38} & \underline{72.72} & \textbf{86.36} \\
\bottomrule\vspace{-2mm}\\
\multicolumn{6}{@{}p{\linewidth}@{}}{\footnotesize 5,000 samples of HaluEval (QA) were used; HaluEval and HDM-Bench contain no training split; HaloScope, LookbackLens, INSIDE, Perplexity, and RAGognizer use Qwen3-4B-Instruct-2507 as proxy generator; HallucinationProbes uses a LoRA-KL probe on LLaMA-3.1-8B ($\lambda_{\text{KL}}=0.05$); \textbf{Bold}: best; \underline{Underline}: second-best; \textsuperscript{\textdagger} same-dataset evaluation (split-disjoint).}
\end{tabularx}
\end{table}

\subsubsection{Response-Level Aggregation}
While RAGognizer operates natively on the token level, many applications and detection approaches use simpler response-level hallucination scores. We therefore use max-pooling to aggregate token-level predictions and report response-level AUROC in Table~\ref{tab:hallucination_auroc_bm_response_level}. With this aggregation, RAGognizer performs quite strongly across QA benchmarks with the highest average AUROC compared to all state-of-the-art methods. 
In the case of sampling-based detectors, we generate five additional responses per prompt and evaluate $100$ randomly sampled prompts per dataset category (for RAGognize, all samples are used). RAGognizer consistently outperforms general NLI-based baselines such as DeBERTa-v3 and even compares favorably with larger or more specialized fact-checking models. HallucinationProbes, as a similar approach, shows good performance as well, in two cases even outperforming RAGognizer, which further strengthens the case for such joint training of LoRA adapters and an integrated detection probe.

\begin{table}[tbp]
\centering
\caption{Response-Level AUROC (\%) on ConflictQA (PopQA) Under Contextual--Parametric Alignment Settings.}
\label{tab:conflictqa_auroc}
\small
\begin{tabular}{@{} l *{4}{>{\centering\arraybackslash}p{1.4cm}} @{}}
\toprule
\textbf{Approach} & \textbf{Ctr-P} & \textbf{P-Alg} & \textbf{NoCtx} & \textbf{All} \\
\midrule
\multicolumn{5}{@{}l}{\textit{Black-box detectors}} \\
MiniCheck-7B & \textbf{95.41} & \textbf{98.14} & 68.49 & \textbf{91.65} \\
HDM2-3B & \underline{94.62} & \underline{96.72} & 58.91 & 87.63 \\
DeBERTa-v3 (Con.) & 91.05 & 96.26 & 61.01 & 72.95 \\
DeBERTa-v3 (Ent.) & 93.36 & 97.37 & 63.54 & 83.73 \\
SelfCheckGPT (NLI) & 84.73 & 93.72 & 66.75 & 82.18 \\
LettuceDetect-L & 93.50 & 96.56 & 30.85 & 74.01 \\
HHEM-2.1-Open & 93.01 & 97.37 & 48.98 & 85.55 \\
\midrule
\multicolumn{5}{@{}l}{\textit{White-box detectors}} \\
Semantic Entropy & 70.20 & 65.56 & 61.58 & 65.12 \\
INSIDE (EigenScore) & 62.52 & 64.78 & 58.83 & 61.20 \\
Perplexity & 82.71 & 81.28 & 61.84 & 74.53 \\
HallucinationProbes & 85.13 & 97.35 & \textbf{72.29} & 85.20 \\
\textbf{RAGognizer (Ours)} & 93.81 & 96.11 & \underline{69.26} & \underline{89.33} \\
\bottomrule\vspace{-2mm}\\
\multicolumn{5}{@{}p{10.5cm}@{}}{\footnotesize \textbf{Ctr-P}: Counter-Parametric; \textbf{P-Alg}: Parametric-Aligned; \textbf{NoCtx}: No Context; INSIDE, Perplexity, and RAGognizer use Qwen3-4B-Instruct-2507 as proxy generator; HallucinationProbes uses a LoRA-KL probe on LLaMA-3.1-8B ($\lambda_{\text{KL}}=0.05$); \textbf{Bold}: best; \underline{Underline}: second-best.}
\end{tabular}
\end{table}

\subsection{Generalization to Detection in Open-Domain Scenarios}
RAGognizer has been trained in a rigorous closed-domain setting for detecting hallucinations with respect to a provided context. More generally in real-world use cases, requests to LLMs often involve both contextual and parametric knowledge. Therefore, we are interested in how hallucination signals learned in the closed-domain setup transfer to the more general case in which contextual evidence may align or contradict the model's parametric knowledge. We, therefore, evaluate RAGognizer on the PopQA subset of ConflictQA~\citep{xie2024adaptivechameleonstubbornsloth}, assuming that information from the most popular Wikipedia articles is contained in the models’ parametric knowledge.

Table~\ref{tab:conflictqa_auroc} reports response-level AUROC under three scenarios that differentiate with respect to how the given context relates to the parametric knowledge: First, these two are aligned (Parametric-Aligned: P-Alg). Second, we provide a context that contradicts parametric knowledge (Counter-Parametric: Ctr-P), and last, there is no context provided at all (NoCtx). Across Parametric-Aligned prompts, most detectors, including RAGognizer, achieve high performance.
In Counter-Parametric scenarios, where retrieved evidence contradicts parametric knowledge, RAGognizer remains competitive, achieving an AUROC of $93.81\%$, close to specialized fact-checking models such as MiniCheck-7B. The No Context condition appears to be particularly difficult for all hallucination detection approaches, as performance becomes much worse. Here, HallucinationProbes achieves the highest AUROC ($72.29\%$), consistent with its training objective that treats parametric knowledge as the primary reference signal. RAGognizer ranks second ($69.26\%$), indicating that hallucination signals learned from closed-domain supervision partially transfer to settings without contextual grounding, but do not fully subsume parametric knowledge-based detection. Overall, in this case, RAGognizer generalizes surprisingly well to open-domain and mixed scenarios. It is only outperformed by MiniCheck-7B as a heavy-weight and nonintegrated detector.

To summarize, these results suggest that closed-domain hallucination signals overlap with---but are not identical to--- open-domain uncertainty or truthfulness signals. While RAGognizer generalizes beyond its training regime, the observed performance differences across scenarios highlight that hallucination detection is still conditioned on the underlying knowledge source to a certain degree. This supports our central premise: distinguishing contextual and parametric knowledge is not merely a conceptual choice, but a practical requirement for understanding and modeling hallucinations in LLMs.

\section{Discussion \& Conclusion}
Our results show that hallucination detection is closely tied to internal representations. Jointly optimizing language modeling and token-level hallucination detection consistently increases the separability of hallucinated and grounded tokens, while substantially reducing hallucination rates during generation without degrading language quality or relevance. In contrast to post-hoc probing or purely logit-based heuristics, integrating hallucination supervision during training resulted in better and transferable detection signals. While trained exclusively in closed-domain settings, RAGognizer generalizes well to mixed and open-domain scenarios, indicating that hallucination-related signals learned with respect to contextual grounding partially transfer to more ambiguous cases. Crucially, this transfer relies on first establishing a clear-cut definition of which information is available to the model when judging hallucinations. The introduced RAGognize dataset provides such a well-defined closed-domain dataset, enabling principled learning before we extended this to scenarios where contextual and parametric knowledge interact.

Beyond accuracy, RAGognizer, in addition, offers practical advantages for deployment as it provides real-time token-level hallucination scores with only 3.7M additional parameters, avoiding the computational overhead of external verification models, but also being bound to the host LLM and to the fine-tuning of it.
As a limitation of the current study, we focused on single-turn interactions and on one task (Q\&A). We relied on automated annotation using Gemini~2.5~Flash, employed a fixed weighting between the interacting loss signals, and did not monitor performance on other non-hallucination benchmarks; these aspects should be further analyzed in future work.

Overall, our findings support the view that hallucination detection is fundamentally a representation learning problem and that integrating detection signals into the training process provides a principled path toward more reliable language models in retrieval-augmented generation.

\section{Ethical Considerations}
The dataset introduced in this work is generated and annotated using LLMs and is derived from recent factual statements sourced from Wikipedia. As a result, it may reflect biases, factual inaccuracies, or stylistic artifacts inherited from the underlying models, and it should not be interpreted as representing factual ground truth. The dataset is intended solely for research purposes, particularly for analyzing model behavior under controlled conditions, and should not be used in high-stakes or real-world decision-making settings.



\bibliographystyle{unsrtnat}  
\bibliography{references}  

\begin{thebibliography}{44}
\providecommand{\natexlab}[1]{#1}
\providecommand{\url}[1]{\texttt{#1}}
\expandafter\ifx\csname urlstyle\endcsname\relax
  \providecommand{\doi}[1]{doi: #1}\else
  \providecommand{\doi}{doi: \begingroup \urlstyle{rm}\Url}\fi

\bibitem[Brown et~al.(2020)Brown, Mann, Ryder, and et~al.]{brown2020}
Tom~B. Brown, Benjamin Mann, Nick Ryder, and et~al.
\newblock Language models are few-shot learners, 2020.
\newblock URL \url{https://arxiv.org/abs/2005.14165}.

\bibitem[Huang et~al.(2025)Huang, Yu, Ma, Zhong, Feng, Wang, Chen, Peng, Feng,
  Qin, and Liu]{huang2025}
Lei Huang, Weijiang Yu, Weitao Ma, Weihong Zhong, Zhangyin Feng, Haotian Wang,
  Qianglong Chen, Weihua Peng, Xiaocheng Feng, Bing Qin, and Ting Liu.
\newblock A survey on hallucination in large language models: Principles,
  taxonomy, challenges, and open questions.
\newblock \emph{ACM Transactions on Information Systems}, 43\penalty0
  (2):\penalty0 1--55, January 2025.
\newblock ISSN 1558-2868.
\newblock \doi{10.1145/3703155}.
\newblock URL \url{http://dx.doi.org/10.1145/3703155}.

\bibitem[Petroni et~al.(2019)Petroni, Rockt{\"a}schel, Lewis, Bakhtin, Wu,
  Miller, and Riedel]{petroni2019languagemodelsknowledgebases}
Fabio Petroni, Tim Rockt{\"a}schel, Patrick Lewis, Anton Bakhtin, Yuxiang Wu,
  Alexander~H. Miller, and Sebastian Riedel.
\newblock Language models as knowledge bases?, 2019.
\newblock URL \url{https://arxiv.org/abs/1909.01066}.

\bibitem[Xu et~al.(2024)Xu, Qi, Guo, Wang, Wang, Zhang, and
  Xu]{xu2024knowledgeconflictsllmssurvey}
Rongwu Xu, Zehan Qi, Zhijiang Guo, Cunxiang Wang, Hongru Wang, Yue Zhang, and
  Wei Xu.
\newblock Knowledge conflicts for {LLMs}: A survey, 2024.
\newblock URL \url{https://arxiv.org/abs/2403.08319}.

\bibitem[Lewis et~al.(2021)Lewis, Perez, Piktus, Petroni, Karpukhin, Goyal,
  K{\"u}ttler, Lewis, tau Yih, Rockt{\"a}schel, Riedel, and
  Kiela]{lewis2021retrievalaugmentedgenerationknowledgeintensivenlp}
Patrick Lewis, Ethan Perez, Aleksandra Piktus, Fabio Petroni, Vladimir
  Karpukhin, Naman Goyal, Heinrich K{\"u}ttler, Mike Lewis, Wen tau Yih, Tim
  Rockt{\"a}schel, Sebastian Riedel, and Douwe Kiela.
\newblock Retrieval-augmented generation for knowledge-intensive {NLP} tasks,
  2021.
\newblock URL \url{https://arxiv.org/abs/2005.11401}.

\bibitem[Agrawal et~al.(2024)Agrawal, Suzgun, Mackey, and
  Kalai]{agrawal2024languagemodelsknowtheyre}
Ayush Agrawal, Mirac Suzgun, Lester Mackey, and Adam~Tauman Kalai.
\newblock Do language models know when they're hallucinating references?, 2024.
\newblock URL \url{https://arxiv.org/abs/2305.18248}.

\bibitem[Niu et~al.(2024)Niu, Wu, Zhu, Xu, Shum, Zhong, Song, and
  Zhang]{niu2024ragtruthhallucinationcorpusdeveloping}
Cheng Niu, Yuanhao Wu, Juno Zhu, Siliang Xu, Kashun Shum, Randy Zhong, Juntong
  Song, and Tong Zhang.
\newblock Ragtruth: A hallucination corpus for developing trustworthy
  retrieval-augmented language models, 2024.
\newblock URL \url{https://arxiv.org/abs/2401.00396}.

\bibitem[Yang et~al.(2025)Yang, Li, Yang, and
  et~al.]{yang2025qwen3technicalreport}
An~Yang, Anfeng Li, Baosong Yang, and et~al.
\newblock Qwen3 technical report, 2025.
\newblock URL \url{https://arxiv.org/abs/2505.09388}.

\bibitem[Jelinek et~al.(1977)Jelinek, Mercer, Bahl, and
  Baker]{Jelinek1977PerplexityaMO}
Frederick Jelinek, Robert~L. Mercer, Lalit~R. Bahl, and Janet~M. Baker.
\newblock Perplexity---a measure of the difficulty of speech recognition tasks.
\newblock \emph{Journal of the Acoustical Society of America}, 62, 1977.
\newblock URL \url{https://api.semanticscholar.org/CorpusID:121680873}.

\bibitem[Farquhar et~al.(2024)Farquhar, Kossen, Kuhn, others, and
  Gal]{Farquhar2024DetectingHallucinations}
Sebastian Farquhar, Jannik Kossen, Lorenz Kuhn, others, and Yarin Gal.
\newblock Detecting hallucinations in large language models using semantic
  entropy.
\newblock \emph{Nature}, 630:\penalty0 625--630, 2024.
\newblock \doi{10.1038/s41586-024-07421-0}.
\newblock URL \url{https://www.nature.com/articles/s41586-024-07421-0}.

\bibitem[Chen et~al.(2024)Chen, Liu, Chen, Gu, Wu, Tao, Fu, and
  Ye]{chen2024insidellmsinternalstates}
Chao Chen, Kai Liu, Ze~Chen, Yi~Gu, Yue Wu, Mingyuan Tao, Zhihang Fu, and
  Jieping Ye.
\newblock {INSIDE}: {LLMs}' internal states retain the power of hallucination
  detection, 2024.
\newblock URL \url{https://arxiv.org/abs/2402.03744}.

\bibitem[Chuang et~al.(2024)Chuang, Qiu, Hsieh, Krishna, Kim, and
  Glass]{chuang2024lookbacklensdetectingmitigating}
Yung-Sung Chuang, Linlu Qiu, Cheng-Yu Hsieh, Ranjay Krishna, Yoon Kim, and
  James Glass.
\newblock Lookback lens: Detecting and mitigating contextual hallucinations in
  large language models using only attention maps, 2024.
\newblock URL \url{https://arxiv.org/abs/2407.07071}.

\bibitem[Azaria and Mitchell(2023)]{azaria2023internalstatellmknows}
Amos Azaria and Tom Mitchell.
\newblock The internal state of an {LLM} knows when it's lying, 2023.
\newblock URL \url{https://arxiv.org/abs/2304.13734}.

\bibitem[Obeso et~al.(2025)Obeso, Arditi, Ferrando, Freeman, Holmes, and
  Nanda]{obeso2025realtimedetectionhallucinatedentities}
Oscar Obeso, Andy Arditi, Javier Ferrando, Joshua Freeman, Cameron Holmes, and
  Neel Nanda.
\newblock Real-time detection of hallucinated entities in long-form generation,
  2025.
\newblock URL \url{https://arxiv.org/abs/2509.03531}.

\bibitem[Hu et~al.(2021)Hu, Shen, Wallis, Allen-Zhu, Li, Wang, Wang, and
  Chen]{hu2021loralowrankadaptationlarge}
Edward~J. Hu, Yelong Shen, Phillip Wallis, Zeyuan Allen-Zhu, Yuanzhi Li, Shean
  Wang, Lu~Wang, and Weizhu Chen.
\newblock {LoRA}: Low-rank adaptation of large language models, 2021.
\newblock URL \url{https://arxiv.org/abs/2106.09685}.

\bibitem[Su et~al.(2024)Su, Wang, Ai, HU, Wu, Zhou, and
  Liu]{su2024unsupervisedrealtimehallucinationdetection}
Weihang Su, Changyue Wang, Qingyao Ai, Yiran HU, Zhijing Wu, Yujia Zhou, and
  Yiqun Liu.
\newblock Unsupervised real-time hallucination detection based on the internal
  states of large language models, 2024.
\newblock URL \url{https://arxiv.org/abs/2403.06448}.

\bibitem[Hu et~al.(2025)Hu, He, Xie, and
  Zhang]{hu2025lrp4ragdetectinghallucinationsretrievalaugmented}
Haichuan Hu, Congqing He, Xiaochen Xie, and Quanjun Zhang.
\newblock {LRP4RAG}: Detecting hallucinations in retrieval-augmented generation
  via layer-wise relevance propagation.
\newblock \emph{arXiv preprint 2408.15533}, 2025.

\bibitem[Zhang et~al.(2025)Zhang, Hu, Zhang, Zhang, and
  Wan]{zhang2025icrprobetrackinghidden}
Zhenliang Zhang, Xinyu Hu, Huixuan Zhang, Junzhe Zhang, and Xiaojun Wan.
\newblock {ICR Probe}: Tracking hidden state dynamics for reliable
  hallucination detection in llms: Tracking hidden state dynamics for reliable
  hallucination detection in {LLMs}, 2025.
\newblock URL \url{https://arxiv.org/abs/2507.16488}.

\bibitem[Manakul et~al.(2023)Manakul, Liusie, and
  Gales]{manakul2023selfcheckgptzeroresourceblackboxhallucination}
Potsawee Manakul, Adian Liusie, and Mark J.~F. Gales.
\newblock {SelfCheckGPT}: Zero-resource black-box hallucination detection for
  generative large language models: Zero-resource black-box hallucination
  detection for generative large language models, 2023.
\newblock URL \url{https://arxiv.org/abs/2303.08896}.

\bibitem[He et~al.(2023)He, Gao, and
  Chen]{he2023debertav3improvingdebertausing}
Pengcheng He, Jianfeng Gao, and Weizhu Chen.
\newblock {DeBERTaV3}: Improving {DeBERTa} using {ELECTRA-Style} pre-training
  with gradient-disentangled embedding sharing, 2023.
\newblock URL \url{https://arxiv.org/abs/2111.09543}.

\bibitem[Tang et~al.(2024)Tang, Laban, and
  Durrett]{tang2024minicheckefficientfactcheckingllms}
Liyan Tang, Philippe Laban, and Greg Durrett.
\newblock {MiniCheck}: Efficient fact-checking of {LLMs} on grounding
  documents, 2024.
\newblock URL \url{https://arxiv.org/abs/2404.10774}.

\bibitem[Ravi et~al.(2024)Ravi, Mielczarek, Kannappan, Kiela, and
  Qian]{ravi2024lynxopensourcehallucination}
Selvan~Sunitha Ravi, Bartosz Mielczarek, Anand Kannappan, Douwe Kiela, and
  Rebecca Qian.
\newblock Lynx: An open source hallucination evaluation model, 2024.
\newblock URL \url{https://arxiv.org/abs/2407.08488}.

\bibitem[Padhi et~al.(2024)Padhi, Nagireddy, Cornacchia, and
  et~al.]{padhi2024graniteguardian}
Inkit Padhi, Manish Nagireddy, Giandomenico Cornacchia, and et~al.
\newblock Granite guardian, 2024.
\newblock URL \url{https://arxiv.org/abs/2412.07724}.

\bibitem[Mendelevitch et~al.(2024)Mendelevitch, Bao, Li, and Luo]{vectara2024}
Ofer Mendelevitch, Forrest Bao, Miaoran Li, and Rogger Luo.
\newblock {HHEM 2.1}: A better hallucination detection model and a new
  leaderboard.
\newblock Vectara blog, Aug 2024.
\newblock URL
  \url{https://www.vectara.com/blog/hhem-2-1-a-better-hallucination-detection-model}.

\bibitem[Es et~al.(2025)Es, James, Espinosa-Anke, and
  Schockaert]{es2025ragasautomatedevaluationretrieval}
Shahul Es, Jithin James, Luis Espinosa-Anke, and Steven Schockaert.
\newblock Ragas: Automated evaluation of retrieval augmented generation, 2025.
\newblock URL \url{https://arxiv.org/abs/2309.15217}.

\bibitem[Paudel et~al.(2025)Paudel, Lyzhov, Joshi, and
  Anand]{paudel2025hallucinothallucinationdetectioncontext}
Bibek Paudel, Alexander Lyzhov, Preetam Joshi, and Puneet Anand.
\newblock {HalluciNot}: Hallucination detection through context and common
  knowledge verification, 2025.
\newblock URL \url{https://arxiv.org/abs/2504.07069}.

\bibitem[Yeh et~al.(2025)Yeh, Li, and
  Mallick]{yeh2025luminadetectinghallucinationsrag}
Samuel Yeh, Sharon Li, and Tanwi Mallick.
\newblock {LUMINA}: Detecting hallucinations in {RAG} system with
  context-knowledge signals, 2025.
\newblock URL \url{https://arxiv.org/abs/2509.21875}.

\bibitem[Li et~al.(2023)Li, Cheng, Zhao, Nie, and
  Wen]{li2023haluevallargescalehallucinationevaluation}
Junyi Li, Xiaoxue Cheng, Wayne~Xin Zhao, Jian-Yun Nie, and Ji-Rong Wen.
\newblock {HaluEval}: A large-scale hallucination evaluation benchmark for
  large language models, 2023.
\newblock URL \url{https://arxiv.org/abs/2305.11747}.

\bibitem[Ridder and
  Schilling(2025)]{ridder2025halluragdatasetdetectingcloseddomain}
Fabian Ridder and Malte Schilling.
\newblock The {HalluRAG} dataset: Detecting closed-domain hallucinations in
  {RAG} applications using an {LLM}'s internal states, 2025.
\newblock URL \url{https://arxiv.org/abs/2412.17056}.

\bibitem[Touvron et~al.(2023)Touvron, Lavril, Izacard, and
  et~al.]{touvron2023llamaopenefficientfoundation}
Hugo Touvron, Thibaut Lavril, Gautier Izacard, and et~al.
\newblock {LLaMA}: Open and efficient foundation language models, 2023.
\newblock URL \url{https://arxiv.org/abs/2302.13971}.

\bibitem[Grattafiori et~al.(2024)Grattafiori, Dubey, Jauhri, and
  et~al.]{grattafiori2024llama3herdmodels}
Aaron Grattafiori, Abhimanyu Dubey, Abhinav Jauhri, and et~al.
\newblock The {Llama} 3 herd of models, 2024.
\newblock URL \url{https://arxiv.org/abs/2407.21783}.

\bibitem[Jiang et~al.(2023)Jiang, Sablayrolles, Mensch, Bamford, Chaplot,
  de~las Casas, Bressand, Lengyel, Lample, Saulnier, Lavaud, Lachaux, Stock,
  Scao, Lavril, Wang, Lacroix, and Sayed]{jiang2023mistral7b}
Albert~Q. Jiang, Alexandre Sablayrolles, Arthur Mensch, Chris Bamford,
  Devendra~Singh Chaplot, Diego de~las Casas, Florian Bressand, Gianna Lengyel,
  Guillaume Lample, Lucile Saulnier, L{\'e}lio~Renard Lavaud, Marie-Anne
  Lachaux, Pierre Stock, Teven~Le Scao, Thibaut Lavril, Thomas Wang,
  Timoth{\'e}e Lacroix, and William~El Sayed.
\newblock Mistral 7b, 2023.
\newblock URL \url{https://arxiv.org/abs/2310.06825}.

\bibitem[Comanici et~al.(2025)Comanici, Bieber, Schaekermann, and
  et~al.]{comanici2025gemini25pushingfrontier}
Gheorghe Comanici, Eric Bieber, Mike Schaekermann, and et~al.
\newblock Gemini 2.5: Pushing the frontier with advanced reasoning,
  multimodality, long context, and next generation agentic capabilities, 2025.
\newblock URL \url{https://arxiv.org/abs/2507.06261}.

\bibitem[Chen et~al.(2023)Chen, Xiao, Zhang, Luo, Lian, and Liu]{bge_m3}
Jianlv Chen, Shitao Xiao, Peitian Zhang, Kun Luo, Defu Lian, and Zheng Liu.
\newblock {BGE M3-Embedding}: Multi-lingual, multi-functionality,
  multi-granularity text embeddings through self-knowledge distillation, 2023.

\bibitem[Wei et~al.(2023)Wei, Wang, Schuurmans, Bosma, Ichter, Xia, Chi, Le,
  and Zhou]{wei2023chainofthoughtpromptingelicitsreasoning}
Jason Wei, Xuezhi Wang, Dale Schuurmans, Maarten Bosma, Brian Ichter, Fei Xia,
  Ed~Chi, Quoc Le, and Denny Zhou.
\newblock Chain-of-thought prompting elicits reasoning in large language
  models, 2023.
\newblock URL \url{https://arxiv.org/abs/2201.11903}.

\bibitem[CH-Wang et~al.(2024)CH-Wang, Durme, Eisner, and
  Kedzie]{chwang2024androidsknowtheyredreaming}
Sky CH-Wang, Benjamin~Van Durme, Jason Eisner, and Chris Kedzie.
\newblock Do androids know they're only dreaming of electric sheep?, 2024.
\newblock URL \url{https://arxiv.org/abs/2312.17249}.

\bibitem[Duan et~al.(2024)Duan, Yang, and
  Tam]{duan2024llmsknowhallucinationempirical}
Hanyu Duan, Yi~Yang, and Kar~Yan Tam.
\newblock Do {LLMs} know about hallucination? an empirical investigation of
  {LLM}'s hidden states, 2024.
\newblock URL \url{https://arxiv.org/abs/2402.09733}.

\bibitem[Xie et~al.(2024)Xie, Zhang, Chen, Lou, and
  Su]{xie2024adaptivechameleonstubbornsloth}
Jian Xie, Kai Zhang, Jiangjie Chen, Renze Lou, and Yu~Su.
\newblock Adaptive chameleon or stubborn sloth: Revealing the behavior of large
  language models in knowledge conflicts, 2024.
\newblock URL \url{https://arxiv.org/abs/2305.13300}.

\bibitem[Kov{\'a}cs and Recski(2025)]{kovacs2025lettucedetect}
{\'A}d{\'a}m Kov{\'a}cs and G{\'a}bor Recski.
\newblock Lettucedetect: A hallucination detection framework for {RAG}
  applications.
\newblock \emph{arXiv preprint arXiv:2502.17125}, 2025.

\bibitem[Du et~al.(2024)Du, Xiao, and
  Li]{du2024haloscopeharnessingunlabeledllm}
Xuefeng Du, Chaowei Xiao, and Yixuan Li.
\newblock {HaloScope}: Harnessing unlabeled {LLM} generations for hallucination
  detection, 2024.
\newblock URL \url{https://arxiv.org/abs/2409.17504}.

\bibitem[Team et~al.(2025)Team, Kamath, and
  et~al.]{gemmateam2025gemma3technicalreport}
Gemma Team, Aishwarya Kamath, and Johan~Ferret et~al.
\newblock Gemma 3 technical report, 2025.
\newblock URL \url{https://arxiv.org/abs/2503.19786}.

\bibitem[Amini et~al.(2025)Amini, Banaszak, Benoit, and
  et~al.]{amini2025lfm2technicalreport}
Alexander Amini, Anna Banaszak, Harold Benoit, and et~al.
\newblock {LFM2} technical report, 2025.
\newblock URL \url{https://arxiv.org/abs/2511.23404}.

\bibitem[{Meta AI}(2024)]{llama32-1b}
{Meta AI}.
\newblock {LLaMA} 3.2 {1B} language model.
\newblock Hugging Face model card,
  \url{https://huggingface.co/meta-llama/Llama-3.2-1B}, 2024.
\newblock [Online; accessed 5-Nov-2025].

\bibitem[{IBM Research}(2025)]{granite2025}
{IBM Research}.
\newblock Granite 4.0 language models.
\newblock GitHub,
  \url{https://github.com/ibm-granite/granite-4.0-language-models}, 2025.
\newblock [Online; accessed 5-Nov-2025].

\end{thebibliography}


\clearpage
\begin{appendices}
\renewcommand{\thefigure}{\Alph{figure}}
\renewcommand{\thetable}{\Alph{table}}
\renewcommand{\thelstlisting}{\Alph{lstlisting}}

\counterwithin*{figure}{section}
\counterwithin*{table}{section}
\counterwithin*{lstlisting}{section}

\section*{Appendix}
\label{sec:appendix}


\begin{table}[h]
\centering
\caption{Comparison of Hallucination Detectors. Lvl.: Detection Granularity (Token/Response); Samp.: Requires Multiple Samples; Backbone: Underlying Model.}

\label{tab:detector_comparison}
\small
\setlength{\tabcolsep}{6pt}
\begin{tabular}{l c c l}
\toprule
\textbf{Approach} & \textbf{Lvl.} & \textbf{Samp.} & \textbf{Backbone}  \\
\midrule
\multicolumn{4}{l}{\textit{Model‑Agnostic (Black‑box)}} \\
LettuceDetect \citep{kovacs2025lettucedetect} & Tok & No & ModernBERT \\
HDM-2 \citep{paudel2025hallucinothallucinationdetectioncontext} & Tok & No & LLM \\
DeBERTa‑based NLI \citep{he2023debertav3improvingdebertausing} & Resp & No & DeBERTa \\
MiniCheck \citep{tang2024minicheckefficientfactcheckingllms} & Resp & No & Llama3 \\
Lynx \citep{ravi2024lynxopensourcehallucination} & Resp & No & Llama3 \\
Granite Guardian 3.3 \citep{padhi2024graniteguardian} & Resp & No & Granite 3.3 \\
HHEM‑2.1-Open \citep{vectara2024} & Resp & No & FLAN‑T5‑base \\
RAGAS Faithfulness \citep{es2025ragasautomatedevaluationretrieval} & Resp & No & LLM \\
SelfCheckGPT \citep{manakul2023selfcheckgptzeroresourceblackboxhallucination} & Resp & Yes & Generator \\
\midrule
\multicolumn{4}{l}{\textit{Model‑Intrinisic (White‑box)}} \\
Semantic Entropy (SE) \citep{Farquhar2024DetectingHallucinations} & Resp & Yes & Generator \\
INSIDE (EigenScore) \citep{chen2024insidellmsinternalstates} & Resp & Yes & Generator \\
ICR Probe \citep{zhang2025icrprobetrackinghidden} & Resp & No & Generator \\
HaloScope \citep{du2024haloscopeharnessingunlabeledllm} & Resp & No & Generator \\
LRP4RAG \citep{hu2025lrp4ragdetectinghallucinationsretrievalaugmented} & Resp & No & Generator \\
Perplexity (PPL) \citep{Jelinek1977PerplexityaMO} & Tok & No & Generator \\
LUMINA \citep{yeh2025luminadetectinghallucinationsrag} & Tok & No & Generator \\
SAPLMA \citep{azaria2023internalstatellmknows} & Tok & No & Generator \\
MIND \citep{su2024unsupervisedrealtimehallucinationdetection} & Tok & No & Generator \\
Lookback Lens \citep{chuang2024lookbacklensdetectingmitigating} & Tok & No & Generator \\
HallucinationProbes \citep{obeso2025realtimedetectionhallucinatedentities} & Tok & No & Generator \\
\bottomrule
\end{tabular}
\end{table}

\begin{figure}[h]
\centerline{\includegraphics[width=0.68\textwidth]{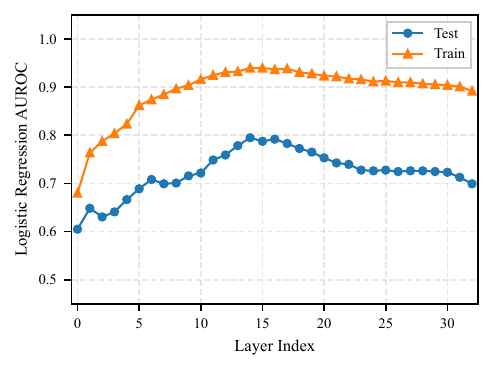}}
\caption{Linear separability of hallucinated vs. grounded tokens across layers of Llama-2-7B-Chat. The plot shows the performance (AUROC) of a Logistic Regression probe trained on the frozen hidden states of each layer to distinguish between grounded and hallucinated tokens.}
\label{fig_auroc_per_layer}
\end{figure}

\begin{table}[hb]
\centering
\caption{SLM Selection and Ablation on RAGognize (Token-Level AUROC after Det. Head FT)}
\label{tab:slm_ablation}
\begin{tabular}{l c c c}
\toprule
\textbf{Model} & \textbf{AUROC (\%)} & \textbf{Layer Depth} & \textbf{LM Head} \\
\midrule
\multicolumn{4}{l}{\textit{SLM Candidates (Head at Mid-Layer)}} \\
Gemma-3-270M \citep{gemmateam2025gemma3technicalreport} & 78.34 & 50\% & \checkmark \\
Qwen3-0.6B \citep{yang2025qwen3technicalreport} & 86.14 & 50\% & \checkmark \\
Gemma-3-1B \citep{gemmateam2025gemma3technicalreport} & 87.33 & 50\% & \checkmark \\
LFM2-1.2B-RAG \citep{amini2025lfm2technicalreport} & 83.33 & 50\% & \checkmark \\
LLaMA-3.2-1B \citep{llama32-1b} & 88.46 & 50\% & \checkmark \\
Granite-4.0-micro \citep{granite2025} & 86.24 & 50\% & \checkmark \\
Qwen3-4B \citep{yang2025qwen3technicalreport} & 92.69 & 50\% & \checkmark \\
\midrule
\multicolumn{4}{l}{\textit{Ablations (Depth \& Head Removal)}} \\
Gemma-3-270M & 81.50 & 50\% & $\times$ \\
Gemma-3-270M & 83.51 & 100\% & $\times$ \\
Qwen3-4B & 93.11 & 50\% & $\times$ \\
Qwen3-4B & 93.68 & 100\% & $\times$ \\
\bottomrule
\multicolumn{4}{p{0.9\columnwidth}}{\footnotesize
\textbf{Layer Depth}: relative transformer depth where the MLP head is injected (50\% = middle layer). \textbf{LM Head}: Language modeling head; \checkmark\ indicates the original LM head is retained, $\times$ indicates removal.}
\end{tabular}
\end{table}

\begin{figure}[h]
\centerline{\includegraphics[width=0.68\textwidth]{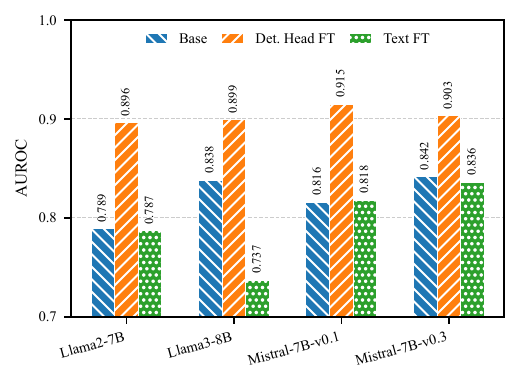}}
\caption{Impact of fine-tuning strategies on hallucination detection: We evaluate the separability of internal states at the middle layer across all response tokens. Det. Head FT results in the highest AUROC, outperforming the Base model. In contrast, standard Text FT often diminishes separability.}
\label{fig_auroc_procedures_hallucinations}
\end{figure}

\end{appendices}

\end{document}